\documentclass{article}

\usepackage[preprint, nonatbib]{neurips_2023}
\usepackage{graphicx}

\usepackage[utf8]{inputenc} % allow utf-8 input
\usepackage[T1]{fontenc}    % use 8-bit T1 fonts
\usepackage{hyperref}       % hyperlinks
\usepackage{url}            % simple URL typesetting
\usepackage{booktabs}       % professional-quality tables
\usepackage{amsfonts}       % blackboard math symbols
\usepackage{nicefrac}       % compact symbols for 1/2, etc.
\usepackage{microtype}      % microtypography
\usepackage{xcolor}         % colors
\usepackage{gensymb}

\title{Towards Automatic Satellite Images Captions Generation Using Large Language Models}

\author{%
  Yingxu He \\
  Department of Computer Science \\
  National University of Singapore \\
  e0139128@u.nus.edu
  \And
  Qiqi Sun  \\  
  College of Life Sciences \\
  Nankai University \\ 
  sunqiqi1018@gmail.com \\
}

\begin{document}

\maketitle

\begin{abstract}
Automatic image captioning is a promising technique for conveying visual information using natural language. It can benefit various tasks in satellite remote sensing, such as environmental monitoring, resource management, disaster management, etc. However, one of the main challenges in this domain is the lack of large-scale image-caption datasets, as they require a lot of human expertise and effort to create. Recent research on large language models (LLMs) has demonstrated their impressive performance in natural language understanding and generation tasks. Nonetheless, most of them cannot handle images (GPT-3.5, Falcon, Claude, etc.), while conventional captioning models pre-trained on general ground-view images often fail to produce detailed and accurate captions for aerial images (BLIP, GIT, CM3, CM3Leon, etc.). To address this problem, we propose a novel approach: Automatic Remote Sensing Image Captioning (ARSIC) to automatically collect captions for remote sensing images by guiding LLMs to describe their object annotations. We also present a benchmark model that adapts the pre-trained generative image2text model (GIT) to generate high-quality captions for remote-sensing images. Our evaluation demonstrates the effectiveness of our approach for collecting captions for remote sensing images. 

Many previous studies have shown that LLMs such as GPT-3.5 and GPT-4 are good at understanding semantics but struggle with numerical data and complex reasoning. To overcome this limitation, ARSIC leverages external APIs to perform simple geographical analysis on images, such as object relations and clustering. We perform clustering on the objects and present the significant geometric relations for LLM to make summarizations. The final output of the LLM is several captions that describe the image, which will be further ranked and shortlisted based on language fluency and consistency with the original image.

We fine-tune a pre-trained generative image2text (GIT) model on 7 thousand and 2 thousand image-caption pairs from the Xview and DOTA datasets, which contain satellite images with bounding box annotations for various objects, such as vehicles, constructions, ships, etc. We evaluate our approach on the RSICD dataset, a benchmark dataset for satellite image captioning with 10,892 images and 31,783 captions annotated by human experts. We remove captions with unseen object types from the training data and obtain 1746 images with more than 5 thousand captions, where we achieve a CIDEr-D score of 85.93, demonstrating the effectiveness and potential of our approach for automatic image captioning in satellite remote sensing. Overall, this work presents a feasible way to guide them to interpret geospatial datasets and generate accurate image captions for training end-to-end image captioning models. Our approach reduces the need for human annotation and can be easily applied to datasets or domains.
\end{abstract}

\clearpage

\section{Introduction}

Satellite remote sensing is essential in numerous fields, such as disaster management, environmental monitoring, and resource management. It involves analyzing images captured from space, focusing on detecting and classifying objects on Earth's surface to produce useful spatial information. As these images can contain a rich amount of data, automatic image captioning has emerged as an efficient method to interpret and convey the visual information in these images using natural language.

Despite its significant potential, a major challenge in automatic image captioning in satellite remote-sensing images is the scarcity of large-scale image-caption datasets. Creating such datasets is labor-intensive and demands significant human expertise. Often, pre-existing models, such as GPT-3.5\cite{IntroducingChatGPT}, Falcon, and Claude, fall short in their applicability as they are not equipped to interpret numerical data or carry out complex reasoning. Similarly, models like BLIP\cite{DBLP:journals/corr/abs-2201-12086}, GIT\cite{wang2022git}, CM3\cite{DBLP:journals/corr/abs-2201-07520}, and CM3Leon\cite{yu2023scaling} that are pre-trained on general ground-view images struggle to generate precise captions for aerial images. These limitations make it challenging to achieve high-quality automatic captioning for remote-sensing images.

To confront this issue, in this study, we propose a novel approach: Automatic Remote Sensing Image Captioning (ARSIC), which leverages both large language models and satellite data to generate high-quality captions for remote sensing images efficiently. Our contributions are threefold. First, we develop several geographical analysis APIs to detect clusters, identify shapes formed by objects, and calculate distances to offer an enhanced understanding of the image. Second, we automate the process of caption collection by guiding large language models to summarize the results from the geographical APIs into captions. This reduces the need for human annotation considerably. Lastly, we provide a benchmark by finetuning a generative image2text (GIT) model on image-caption pairs collected following our ARSIC approach from the Xview\cite{DBLP:journals/corr/abs-1802-07856} and DOTA\cite{9560031} datasets and tailored to generate high-quality and accurate captions for aerial images. 

The effectiveness of our approach is validated through rigorous testing on the RSICD\cite{lu2017exploring} test dataset, setting a new benchmark CIDEr-D\cite{vedantam2015cider} score in the field. In summary, our work presents an innovative approach towards interpreting and captioning remote sensing images - a method that is not only promising for optimizing end-to-end image captioning models but also flexible enough to be applied across datasets or domains.

\section{Methodology}

In this section, we describe our proposed approach to automatically collect captions for remote sensing images by guiding LLMs to describe their object annotations. In this work, we limit the number of objects in each image to no more than 15, which ensures a relatively simple spatial layout for the LLM. Our approach consists of three main steps: (1) develop APIs to conduct geographical analysis and describe spatial relationships between objects, (2) prompt the API to generate captions with the help from APIs, and (3) caption evaluation and selection. We explain each step in detail below.

\subsection{Spatial Relationship APIs}\label{sec:api}

LLM is incompetent at processing 2-dimensional geographical information, so we implemented several analytical approaches to analyze the spatial relations between objects. Inspired by the captions provided by the RSICD paper, we only focused on analyzing the distances between objects, the concentration of object locations, shapes formed by groups of objects, and significant relations between objects. 

\subsubsection{Distance}

In the Xview and Dota datasets, the size of objects varies a lot. Therefore, using the distance between centers is inappropriate for the distances between objects. For instance, although the centers of two large buildings might be quite far apart, their inner-facing walls might be only a few steps away. Therefore, we consider the shortest distances between bounding boxes as their distance. For the distance between two groups of objects, we represent it with the distance between their closest element, which is normally referred to as the Single Linkage measure in the field of clustering.

\subsubsection{Clustering}\label{sec:clustering}
One of the most important features captured by human eyes is the concentration of objects based on their locations and types, e.g., one tends to easily differentiate a vehicle running on a highway from several buildings standing by the road. On the other hand, people also tend to pay attention to the objects' closest neighbor, e.g., a passenger car next to a truck is easier to draw people's attention than a building relatively further away from the truck. Traditional machine learning clustering algorithms include distance-based algorithms such as K-Means and hierarchical clustering, and density-based clustering such as DBSCAN and its variants. However, the K-Means algorithm often fails to separate outliers from concentrated objects, while the benefits of density-based clustering might be buried in this case, where each image only contains fewer than ten objects. 

\begin{figure*}[h]
\centering
\includegraphics[scale=0.5]{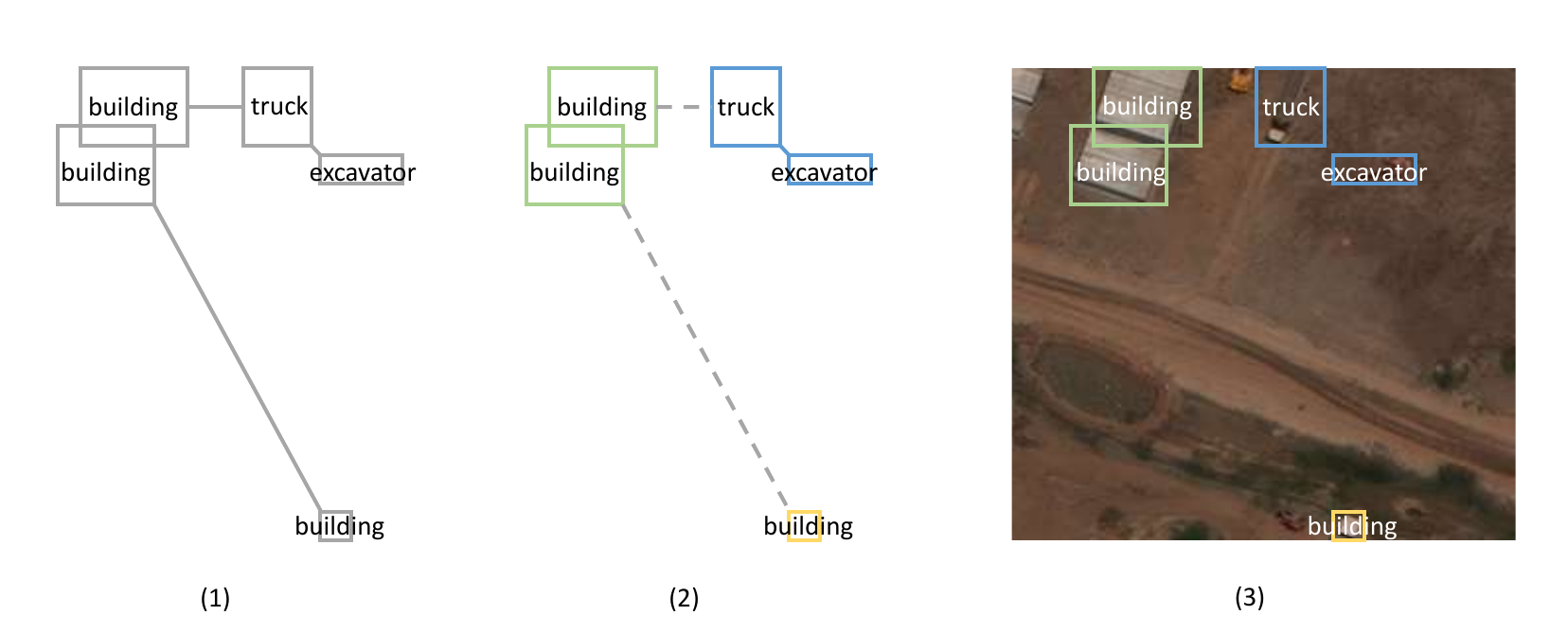}
\vspace{-0.1cm}
\caption{Illustration of the MST-based clustering algorithm. Figure (1) demonstrates the created graph representing the minimal spanning tree. Extra length is added to the distance between objects of different types. Figure (2) shows the clusters formed by cutting long edges. Figure (3) projects the location of the objects to the real image.}
\label{fig:clustering}
\vspace{-0.1cm}
\end{figure*}

In this work, We used the Minimum Spanning Tree (MST) algorithm to connect all the objects in the image and form clusters by removing significantly long edges from the graph. Kruskal's MST algorithm\cite{10.5555/1051910} considers objects' nearest neighbors and simultaneously skips negligible connections, ensuring every tree edge is aligned to humans' observing behavior. We set the threshold at the 75 percentile of the edge weights from the entire dataset. Edges above this threshold were removed from the graph to form clusters, minimizing intra-cluster and maximizing inter-cluster distances. To encourage grouping objects of the same type into the same cluster, We add extra length to distances between objects of different types. Figure \ref{fig:clustering} gives a detailed illustration on the MST-based clustering algorithm. This approach could precisely split objects by type, location, and proximity, which benefits the subsequent geographical analysis. 

\subsubsection{Geometric Shape}\label{sec:shape}

Inspired by the captions provided in the RSICD dataset, the line shape is considered the fundamental shape to be detected in this work. It seems most attractive to human eyes and the basic element of many other complicated shapes. For instance, the square grid street pattern is one of the most popular street patterns used in cities, where lines of buildings are the most fundamental elements. Undeniably, other shapes could also easily draw people's attention, such as circles and squares. Nonetheless, in the setting of this work, where each image contains at most 15 objects, they are less obvious and more difficult to detect. Therefore, we only implemented a method to detect line shapes from groups of objects by inspecting whether the lines formed by corners of bounding boxes are parallel. 

\subsubsection{Geometric Relation}

We review some relations listed in the RSICD paper\cite{lu2017exploring} and come out with our list of relations to be included in the image captions: "stands alone", "near", "in a row", "surrounded by", "between", and "in two sides of". We modified the "in rows" relation from RSICD paper to "in a row", as objects in different rows can be clustered into different groups as is described in section \ref{sec:clustering}, and any possible line shape will be detected by the shape identification algorithm described in section \ref{sec:shape}. Additionally, we propose a "between" relation as the flip side of "in two sides of" to differentiate the situation where there are only objects on the two sides of others from objects circling others 360\degree. In this work, the approaches described above can address relations "stands alone", "near", and "in a row". The relation "surrounded by" is only considered when certain objects are located within the border of another group of objects. The detailed function is achieved by drawing links from the boxes in the middle to the outer ones and calculating the angles between them. The implementation of relations "between" and "in two sides of" are left for future work. 

\subsection{LLM Prompting}

The second step of our approach is to use prompts to guide the LLM to produce a caption following a similar pattern. With the APIs implemented in section \ref{sec:api}, there are many options to prompt the LLM and guide it to generate the ideal captions. Following the recently popular idea of treating the LLMs as a controller or action dispatcher\cite{zhang2023datacopilot}, one approach could be allowing the language model to plan its actions and execute the functions in sequences to obtain helpful geographical analysis results. For instance, the recently developed ReAct\cite{yao2023react} approach synergizes the reasoning and executing process of LLM to enhance its capability of handling complex tasks. It allows great flexibility in geographical analysis and greater diversity in the generated captions. Nonetheless, the LLM tends to experience difficulty discovering eye-catching geographical relations and is easily flooded with less important information received during the action execution process. 

To solve the problem, we adopted the advantage of the MST algorithm, which reveals the most important neighbors for both clusters and stand-alone objects, from where we can easily extract the significant geographical relations.  More specifically, we list the presence of every group in each image with their combination and shapes detected, together with stand-alone objects. The significant geometric relations between the boxes are then provided to give the LLM a sense of their spatial relations. In this case, we only present the edges removed during the clustering step (section \ref{sec:clustering}) that connects clusters and stand-alone objects. An illustration of the spatial relations presented and captions created by LLM is provided in figure \ref{fig:prompting_flow}.

\subsubsection{Captions Diversification}

\begin{figure*}[h]
\centering
\includegraphics[scale=0.5]{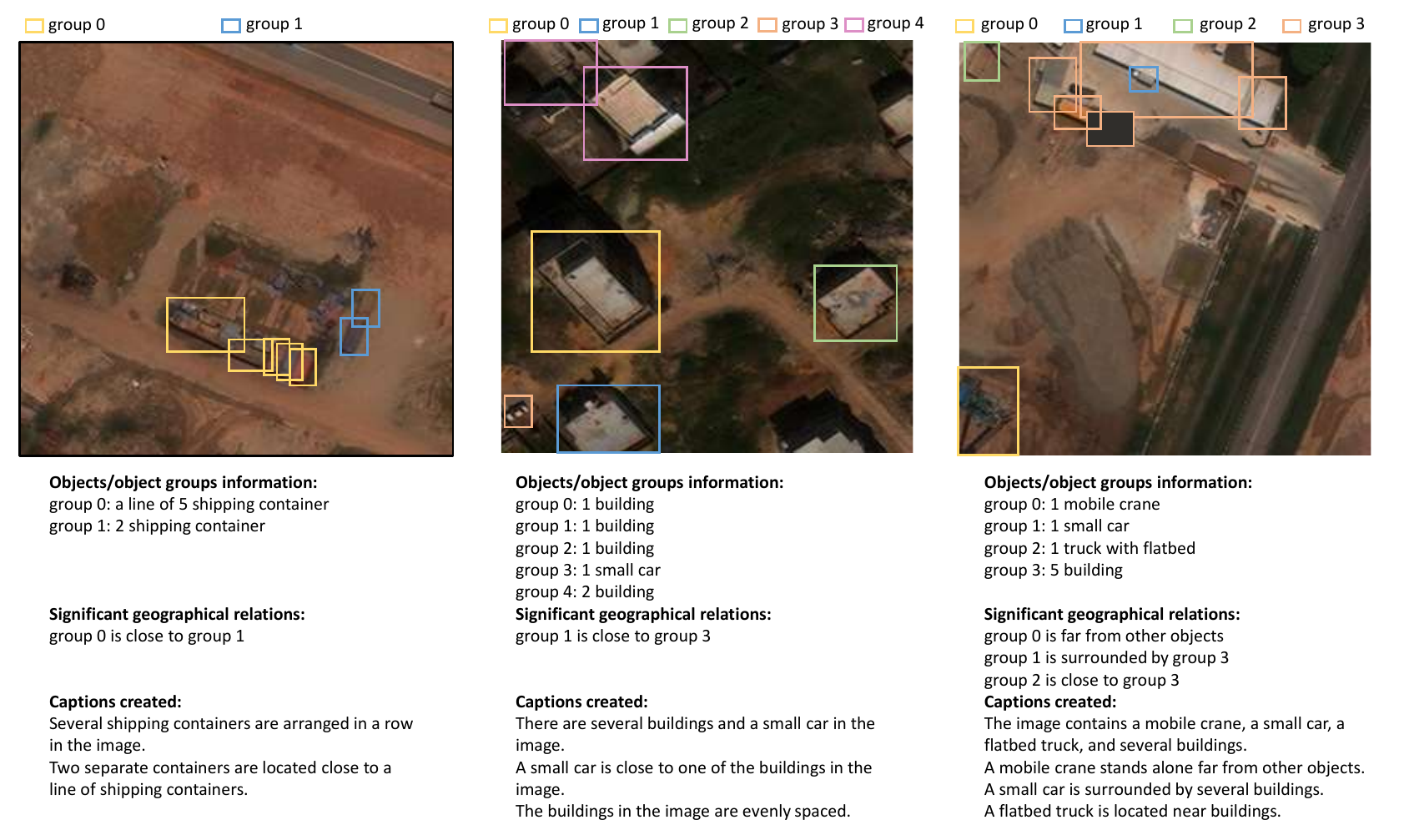}
\vspace{-0.1cm}
\caption{Examples of geographical analysis information and captions generated by LLM. For each example, object information and geographical patterns are provided by our implemented APIs and given to the LLM as input.}
\label{fig:prompting_flow}
\vspace{-0.1cm}
\end{figure*}

Although the prompt already provided necessary clustering information and spatial relations between objects, LLM is not supposed only to bring the clustering information into the spatial relations and create captions, which can be already done by a template-based or rule-based method. The most important role played by LLM is to understand the current spatial layout and paraphrase the potentially redundant or insignificant relations into appropriate captions. For instance, in figure \ref{fig:prompting_flow} (2), the MST-based algorithm detects one building is closer to some buildings than others. However, as the whole image is occupied with different buildings, a caption repeating that relation might bring confusion and ambiguity to the downstream deep-learning models and even human readers. In this case, LLM plays a vital role in evaluating the significance of each spatial relation and performing necessary paraphrasing. 

In this work, the summarising behavior of LLM is ensured by providing necessary examples in the prompt, which is more frequently referred to as the "Few-Shot" prompting technique. We provided several examples where LLM is supposed to synergize the clustering results with the spatial relations to create captions in its own words. Other prompting techniques could potentially achieve the same goal, such as adding descriptions for the expected behaviors or breaking down the reasoning process using Chain of Thought or Tree of Thought techniques. Nonetheless, given the input and expected output format are already complicated, these prompting strategies could bring much more complexity and difficulty into the prompt writing process. Moreover, our experimental results show that few-shot prompting performs more stably than any of the above-mentioned techniques. 

\subsubsection{Response Formatting}

\begin{figure*}[h]
\centering
\includegraphics[scale=0.6]{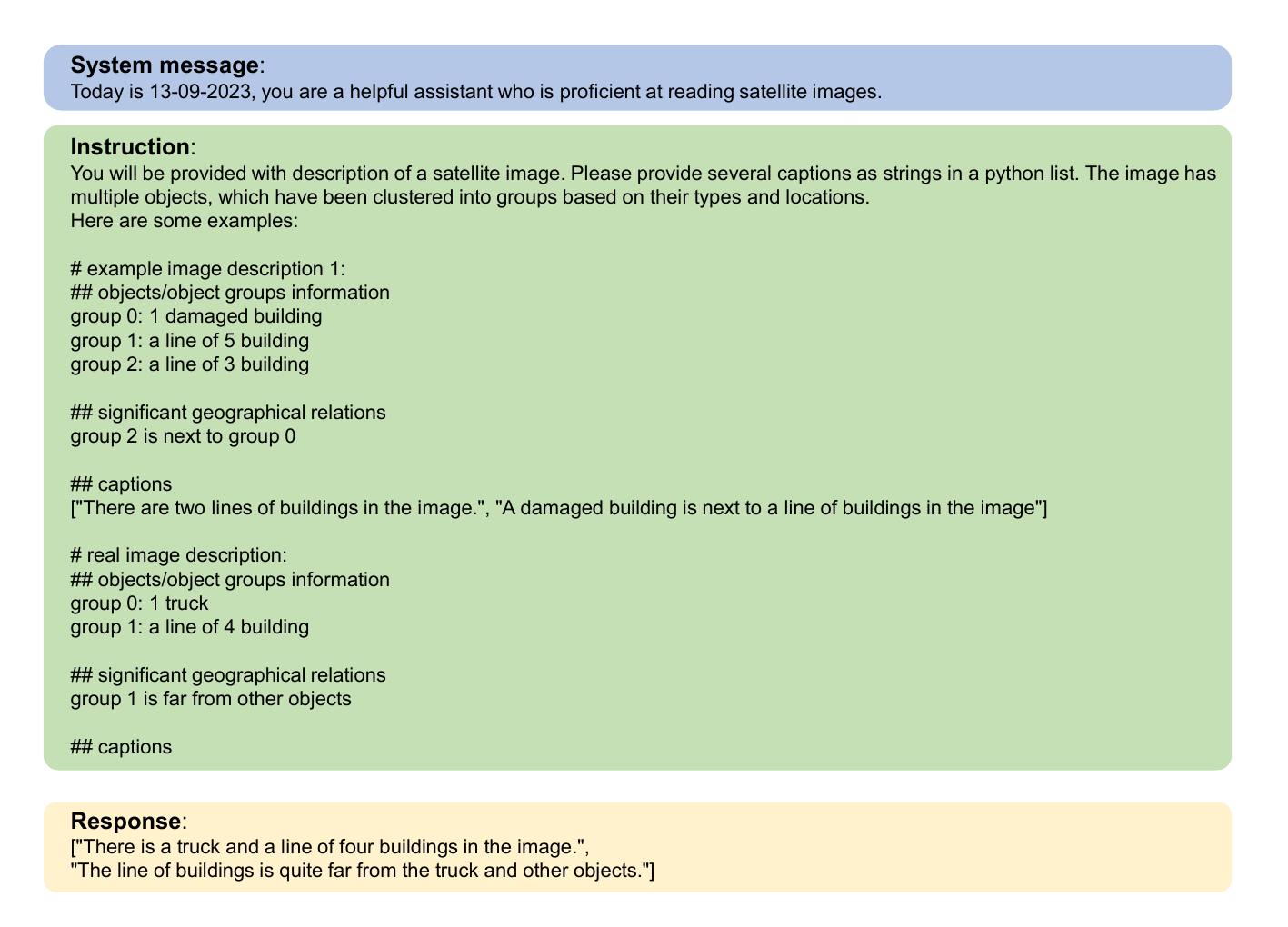}
\vspace{-0.1cm}
\caption{Illustration of the structure of our prompt and output from LLM. More examples are given to the LLM in the prompt, while only one is included here for demonstration.}
\label{fig:prompting}
\vspace{-0.1cm}
\end{figure*}

Additionally, to effectively restrict the response to a computer-readable format, we explicitly instruct the LLM to output the captions in the format of a Python list, whose detailed information has already been included in LLM's pre-training corpus and well embedded in its parametric memory, rather than other customized format that requires extra explanation. It is desired not to have any id of the object groups in the LLM response, which is achieved again by providing examples in the prompt, as introduced in the prior section. It has been stated in many recent research works that few-shot prompting works better than zero-show prompting with prolonged instructions\cite{ye2022unreliability}. The detailed procedures can be shown in figure \ref{fig:prompting}.

\subsection{Caption Evaluation and Selection}

The third step of our approach is to evaluate and select the best caption for each image. We use two criteria to assess the quality of captions: (a) caption quality, which measures how well the caption matches the ground truth annotation, and (b) caption diversity, which measures how different the caption is from other captions generated from other images. We use the following procedure:

\begin{itemize}
\item We filter out captions containing undesirable keywords such as the group's id, like "group 0" or the group's order, like "the first group", which could lead to confusion. 
\item We use pre-trained CLIP to compute a score for each caption based on its match the input image. The evaluator is trained on a large-scale image-caption dataset that covers various domains and scenarios.
\item We use a similarity measure to compute a score for each caption based on caption diversity. The similarity measure compares each caption with captions generated from other images to avoid descriptions that are too vague and broad. 
\item We combine both scores using a weighted average formula to obtain a final score for each caption.
\item We select the caption with the highest final score as the best caption for each image.
\end{itemize}

\clearpage
\bibliographystyle{plain}
\bibliography{refs}

\end{document}